\documentclass[10pt, a4paper]{article}

\usepackage[final]{lrec2026} % this is the new style
\usepackage{tipa}
\usepackage{amsmath} 

\title{A Sociophonetic Analysis of Racial Bias in Commercial ASR Systems Using the Pacific Northwest English Corpus}

\name{Michael Scott, Siyu Liang, Alicia Wassink, Gina-Anne Levow}

\address{
    Department of Linguistics, University of Washington \\
    michael.kelly.scott@gmail.com, \{liangsy, wassink, levow\}@uw.edu
}

\abstract{
This paper presents a systematic evaluation of racial bias in four major commercial automatic speech recognition (ASR) systems using the Pacific Northwest English (PNWE) corpus. We analyze transcription accuracy across speakers from four ethnic backgrounds (African American, Caucasian American, ChicanX, and Yakama) and examine how sociophonetic variation contributes to differential system performance. We introduce a heuristically-determined Phonetic Error Rate (PER) metric that links recognition errors to specific linguistically motivated variables derived from sociophonetic annotation. Our analysis of eleven sociophonetic features reveals that vowel quality variation, particularly resistance to the low-back merger and pre-nasal merger patterns, is systematically associated with differential error rates across ethnic groups, with the most pronounced effects for African American speakers across all evaluated systems. These findings demonstrate that acoustic modeling of dialectal phonetic variation, rather than lexical or syntactic factors, remains a primary source of bias in commercial ASR systems. The study establishes the PNWE corpus as a valuable resource for bias evaluation in speech technologies and provides actionable guidance for improving ASR performance through targeted representation of sociophonetic diversity in training data.
\\ \newline  }

\begin{document}

\maketitleabstract

\section{Introduction}

Automatic Speech Recognition (ASR) systems demonstrate significantly higher error rates for speakers from marginalized communities. \citet{koenecke_racial_2020} found that commercial ASR systems exhibit average word error rates of 35\% for African American speakers compared to 19\% for Caucasian American speakers across five major commercial systems. This disparity represents a critical fairness issue as speech technologies become increasingly ubiquitous in everyday applications, from virtual assistants to automated closed captioning, medical dictation, and educational technology \citep{cunningham_understanding_2024}.

While these disparities are well-documented across multiple studies \citep{tatman_gender_2017, tatman_effects_2017, martin_bias_2023}, the underlying linguistic mechanisms remain underexplored. \citet{wassink_uneven_2022} provided crucial evidence that phonetic rather than lexical or syntactic factors drive these differences, showing that ASR systems exhibit higher error rates for African American English even when lexical content is controlled. This finding suggests that acoustic modeling, rather than language modeling, is the primary source of bias — a hypothesis further supported by subsequent investigations \citep{chang_self-supervised_2024, johnson_exploratory_2024}.

Recent work has begun to address these disparities through various approaches. \citet{dorn_dialect-specific_2019} explored dialect-specific models and targeted training strategies, while others like \citet{zhang_google_2023, wang_can_2024, li_reexamining_2024} have investigated how large-scale multilingual models and self-supervised learning systems handle dialectal variation. Studies using diverse corpora such as AfriSpeech-200 have demonstrated the potential to improve ASR performance across African accents through better data representation \citeplanguageresource{AfriSpeech}. However, these studies often focus on model architectures or training strategies rather than systematically identifying which specific phonetic features of the underrepresented dialect contribute to transcription errors. Understanding these linguistic mechanisms is essential for developing effective mitigation strategies.

This paper contributes a systematic evaluation methodology for measuring ASR bias using the Pacific Northwest English (PNWE) corpus \citeplanguageresource{PNWE}, focusing on eleven sociophonetic variables that exhibit variation across the ethnic communities represented in the corpus. We evaluate four major commercial ASR systems and demonstrate that specific phonetic features, particularly resistance to the low-back merger and variation in pre-nasal /æ/ patterns, are disproportionately associated with transcription errors for African American speakers. Our approach extends existing ASR evaluation frameworks by incorporating detailed sociolinguistic analysis, providing both traditional word error rates and a novel heuristically-determined phonetic error rate that links errors to specific phonological features documented in the sociolinguistic literature \citep{thomas_phonological_2007, green_african_2002}. By identifying the specific phonetic characteristics to evaluate current ASR systems, this work provides actionable guidance for improving system performance and emphasizes the PNWE corpus as a valuable resource for bias evaluation in speech technologies.

\section{Related Work}

\subsection{Racial Disparities in Commercial ASR Systems}

The work of \citet{koenecke_racial_2020} demonstrated substantial racial disparities in commercial ASR systems, finding almost double average word error rates for African American speakers compared to Caucasian American speakers across five major systems. \citet{tatman_gender_2017, tatman_effects_2017} further confirmed these disparities across diverse datasets and speech contexts. These errors have real-world impacts, with users often feeling ``othered'' by technology that fails to recognize their speech \citep{cunningham_understanding_2024}. However, most prior work had focused on documenting disparities rather than identifying specific linguistic mechanisms driving differential performance.

\subsection{Linguistic Factors in ASR Performance Disparities}

\citet{wassink_uneven_2022} provided crucial evidence that phonetic rather than lexical or syntactic factors drive performance differences for African American English speakers, showing that ASR systems exhibit higher error rates for African American speech even when lexical content is controlled. This finding clarifies that the disparity arises primarily from acoustic modeling rather than from language modeling, as text-based models are not the focus of this comparison. Taken together with the findings of \citet{koenecke_racial_2020}, this implicates acoustic modeling as the primary source of bias. Sociolinguistic research has extensively documented systematic phonetic variation in AAL \citep{thomas_phonological_2007, green_african_2002}, including variable low-back merger, pre-nasal /æ/ raising, /ay/ monophthongization, and consonant reduction -- features that create mismatches with models trained on Mainstream American English.

Recent investigations of newer architectures reveal persistent challenges. \citet{chang_self-supervised_2024} showed that self-supervised speech representations continue to struggle with AAL, while \citet{li_reexamining_2024} highlighted confounding factors like audio quality in performance assessments. Despite advances in large-scale multilingual systems \citep{zhang_google_2023} and dialect-specific models \citep{dorn_dialect-specific_2019}, few studies systematically connect specific sociophonetic variables to ASR errors. Our work addresses this gap through the quantitative analysis of eleven phonetic features across multiple commercial systems.

\subsection{Resources for ASR Bias Evaluation}

Evaluating ASR fairness requires corpora with demographically diverse speakers and controlled recording conditions. The Corpus of Regional African American Language \citeplanguageresource{CORAAL} provides extensive AAL recordings, although the audio quality varies. Mozilla Common Voice \citeplanguageresource{CommonVoice} includes demographic metadata and has recently begun adding spontaneous speech, though the release used here primarily contains read prompts that may miss naturalistic variation. The Speech Accent Archive \citeplanguageresource{SpeechAccent} focuses on L2 English, while the Artie Bias Corpus \citeplanguageresource{ArtieBias} has limited sample sizes. AfriSpeech-200 \citeplanguageresource{AfriSpeech} offers 200 hours across 120 African accents but targets clinical domains rather than sociolinguistic variation.

The Pacific Northwest English corpus \citeplanguageresource{PNWE} offers distinct advantages: balanced sampling across five ethnic groups within one region, sociolinguistic interviews eliciting multiple speech styles, consistent high-quality recordings, detailed sociophonetic annotation, and sufficient sample sizes for statistical analysis. Our work validates PNWE for bias evaluation while demonstrating how to connect performance disparities to specific linguistic features, advancing sociolinguistic approaches to fair speech technology \citep{johnson_exploratory_2024, martin_bias_2023}.

\section{Data}

\subsection{The Pacific Northwest English Corpus}

The \textbf{Pacific Northwest English (PNWE)} corpus \citeplanguageresource{PNWE} was designed to document regional and ethnic variation in the Pacific Northwest of the United States, encompassing linguistic and social diversity across co-regional communities in Washington State. All speakers provided informed consent and were compensated for their participation.

The PNWE corpus includes multiple tasks designed to elicit different speech styles, including:
\begin{itemize}
    \item \textbf{Word List (WL)} task: controlled production of lexical items embedded in the carrier phrase “write \_\_\_ today,” targeting specific phonetic contrasts;
    \item \textbf{Lexical Game (LEX)} task: semi-controlled elicitation with variable speech rate and style;
    \item \textbf{Conversational Speech (CS)} task: open-ended sociolinguistic interviews.
\end{itemize}

The complete corpus comprises recordings from 112 speakers across five ethnic groups: African American, Caucasian American, Asian American, ChicanX, and Yakama. All speakers were long-term residents of the Pacific Northwest, native English speakers, and self-identified members of their respective ethnic communities. The corpus includes 73 female and 39 male speakers, with ages ranging from 20 to 93 years. Table~\ref{tab:demographics} presents the detailed demographic composition of the corpus. While the corpus includes balanced representation across ethnic groups for some communities, sample sizes vary, with Caucasian American speakers comprising the majority (n=68) and smaller samples from other groups, particularly African American speakers (n=5). This distribution reflects both the regional demographics of the Pacific Northwest and the challenges of recruiting participants from smaller ethnic communities in the region.

Each recording was transcribed using CLOx \citet{wassink_clox_2020}, a wrapper based on Microsoft's ASR system \citep{microsoft_azure_2022}, with all transcripts subsequently hand-checked and manually segmented by trained linguists in Praat TextGrids \citep{boersma_praat_2007}. The final annotations include orthographic and broad phonetic transcriptions. A custom extension of the CMU Pronouncing Dictionary \citeplanguageresource{CMUDict}, termed PNWEdict, was created for the PNWE data to handle regionally distinctive variants. Metadata for each speaker include age, gender, ethnic identity, and speech task.

\begin{table}[h]
\centering
\small
\begin{tabular}{lcccc}
\hline
\textbf{Ethnicity} & \textbf{Spkrs} & \textbf{F} & \textbf{M} & \textbf{Age} \\
\hline
African Am. & 5 & 3 & 2 & 35–72 \\
Caucasian Am. & 68 & 46 & 22 & 20–93 \\
Asian Am. & 19 & 10 & 9 & 20–93 \\
ChicanX & 12 & 10 & 2 & 27–52 \\
Yakama & 8 & 4 & 4 & 31–59 \\
\hline
\textbf{Total} & 112 & 73 & 39 & 20–93 \\
\hline
\end{tabular}
\caption{Demographic composition of the PNWE corpus.}
\label{tab:demographics}
\end{table}

\subsection{Experiment Dataset}

For our experiments, only the WL task was used due to its consistent lexical content and ease of phonetic alignment. The evaluation subset used in this study includes recordings from 16 speakers balanced across four ethnic groups: African American (AA), Caucasian American (CA), ChicanX (CX), and Yakama (YA). Each group contributed four speakers (two male, two female) for analysis. Speaker ages ranged from 20 to 65 years old (mean \(\approx\) 40). This sampling design allows controlled comparison across ethnic groups while holding regional and recording conditions constant.

\begin{table}[h]
\centering
\small
\begin{tabular}{lcccc}
\hline
\textbf{Ethnicity} & \textbf{Spkrs} & \textbf{F} & \textbf{M} & \textbf{Age} \\
\hline
African Am. & 4 & 2 & 2 & 23–64 \\
Caucasian Am. & 4 & 2 & 2 & 21–61 \\
ChicanX & 4 & 2 & 2 & 22–53 \\
Yakama & 4 & 2 & 2 & 24–59 \\
\hline
\textbf{Total} & 16 & 8 & 8 & 21–64 \\
\hline
\end{tabular}
\caption{Demographic composition of the PNWE evaluation subset in this study.}
\label{tab:demographics}
\end{table}

\section{Methodology}

\subsection{Sociophonetic Variables}

To provide a sociolinguistically informed evaluation of the speech models, eleven sociophonetic variables known to exhibit variation across ethnic and regional varieties of American English \citep{wassink_uneven_2022} were targeted, listed in Table \ref{tab:phonetic_features}. These variables include features documented in African American Language (AAL), Chicano English, and Indigenous varieties, as well as features characteristic of Pacific Northwest regional dialects. These were used to test whether specific features co-occur with transcription errors across ASR systems.

In our test data, each sociolinguistic variable was annotated for contextual occurrence (presence/absence) and measured acoustically where applicable. The annotated TextGrids were then used to cross-reference ASR transcription errors, enabling correlation of linguistic variables with system performance. 

\begin{table*}[t]
\centering
\footnotesize
\setlength{\tabcolsep}{3pt}
\begin{tabular}{p{4cm}p{1.2cm}p{4cm}p{3.2cm}p{2.2cm}}
\hline
\textbf{Feature} & \textbf{Code} & \textbf{Description} & \textbf{Realization} & \textbf{Example} \\
\hline
Low-back merger resistance & \texttt{-AO} & \textipa{/A/}--\textipa{/O/} distinction & Often maintained & \textit{cot--caught} \\
Pre-nasal merger & \texttt{IN} & \textipa{/I/}--\textipa{/E/} merger before nasals & Frequent & \textit{pin--pen} \\
\textipa{/aI/} monophthongization & \texttt{AY} & \textipa{/aI/} glide reduction & \textipa{[a:]} before voiced coda & \textit{time} $\rightarrow$ \textipa{[ta:m]} \\
R-deletion & \texttt{R} & Post-vocalic \textipa{/r/} weakening & \textipa{[a@]}, \textipa{[V]} outcomes & \textit{car} $\rightarrow$ \textipa{[ka@]} \\
\textipa{/T/, /D/} stopping & \texttt{TH-s} & Dental fricative $\rightarrow$ stop & \textipa{[d]}, \textipa{[t]} substitutions & \textit{this} $\rightarrow$ \textipa{[dIs]} \\
\textipa{/T/, /D/} fronting & \texttt{TH-f} & Dental fricative $\rightarrow$ labiodental & \textipa{[f]}, \textipa{[v]} substitutions & \textit{with} $\rightarrow$ \textipa{[wIf]} \\
Consonant cluster reduction & \texttt{CC} & Final cluster simplification & Frequent & \textit{test} $\rightarrow$ \textipa{[tEs]} \\
Word-final devoicing & \texttt{Dv} & Voiced obstruent $\rightarrow$ voiceless & Stylistically variable & \textit{had} $\rightarrow$ \textipa{[h\ae t]} \\
Word-final debuccalization & \texttt{Db} & \textipa{/t/}, \textipa{/d/} $\rightarrow$ \textipa{[P]} or $\emptyset$ & Variable realization & \textit{side} $\rightarrow$ \textipa{[saIP]} \\
Pre-lateral back merger & \texttt{prel-o} & \textipa{/U/}--\textipa{/u/}, \textipa{/2/}--\textipa{/o/} before \textipa{/l/} & Variable & \textit{fool--full} \\
Pre-lateral front merger & \texttt{prel-i} & \textipa{/I/}--\textipa{/i/} merger before \textipa{/l/} & Variable & \textit{feel--fill} \\
\hline
\end{tabular}
\caption{The eleven sociophonetic variables annotated in the PNWE corpus.}
\label{tab:phonetic_features}
\end{table*}

% the last two are first cited as inter-mountain west features (utah) and spread to the west as a results, thus western features. 
% th stopping is more common in the yakama 

\subsection{ASR Systems Evaluated}

We evaluated four major commercial ASR systems that represent the current state-of-the-art in speech recognition technology and are widely deployed in consumer and enterprise applications: Amazon Transcribe \citep{amazon_amazon_2022}, Google Cloud Speech-to-Text \citep{google_google_2022}, Apple Speech \citep{apple_apple_2022}, and IBM Watson Speech to Text \citep{ibm_ibm_2022}.

We used default configuration settings with US English language models and no custom vocabulary or adaptation to evaluate the systems as they would perform for typical users without specialized tuning. 

It is important to note that commercial ASR systems are continuously updated, and performance characteristics may change over time. The results presented here reflect system performance at the time of evaluation in 2022 and should be interpreted as a snapshot of these systems' capabilities rather than permanent characterizations. However, the patterns of phonetic errors identified may remain relevant for understanding ongoing challenges in ASR performance across dialects.

\subsection{Data Preparation}

Approximately 4 hours of speech from the WL task were used in the primary analysis. Speech segments were exported as 16 kHz WAV files. The analysis excluded the LEX and CS tasks because they lacked consistent time-aligned phonetic annotation at the time of the study. However, we expect that extending the same methods to CS data (around 9 additional hours) would likely yield richer results, given the higher frequency of sociophonetic features in conversational speech. 

\subsection{Metrics}

\subsubsection{Word Error Rate (WER)}

Word Error Rate (WER) is the standard metric for ASR evaluation, calculated as the edit distance between the hypothesis transcription and the reference transcription, normalized by the number of words in the reference:

\begin{equation}
\text{WER} = \frac{S + D + I}{N}
\end{equation}

where $S$ is the number of substitutions, $D$ is the number of deletions, $I$ is the number of insertions, and $N$ is the total number of words in the reference transcription.

We computed WER using the \texttt{nist sclite} toolkit \citep{nist_sctk_2021}, which performs optimal string alignment between hypothesis and reference transcriptions. The alignment process accounts for minor variations in word boundaries and punctuation while identifying genuine recognition errors.

WER provides an aggregate measure of transcription accuracy but does not distinguish between different types of errors or identify the linguistic sources of errors. To address this limitation, we complement WER analysis with an additional phonetic error rate metric.

\subsubsection{Phonetic Error Rate (PER)}

To investigate whether phonetic variation contributes to transcription errors, we developed an heuristically-determined Phonetic Error Rate (PER) metric. This metric compares the phonetic realization of what speakers actually said (from broad phonetic transcriptions in the given TextGrids) against the canonical phonetic representation of what the ASR system hypothesized (derived from PNWEdict).

The PER calculation involves the following steps:

\textbf{Step 1: Extract Reference Phonetic Transcriptions}

We extracted phonetic transcriptions from the Praat TextGrids provided in the PNWE corpus, which were manually created by trained linguists during corpus annotation. These transcriptions capture the actual phonetic realizations produced by speakers, including dialectal variations.

\textbf{Step 2: Generate Hypothesis Phonetic Transcriptions}

For each word in the ASR system's orthographic hypothesis, we looked up its canonical pronunciation in PNWEdict, a regionally adapted version of CMUdict providing ARPABET-encoded phonemic representations. Words not found in PNWEdict were generated using standard phonological rules for English inflections (e.g., possessives, plurals, gerunds).

To illustrate the alignment process, Table~\ref{tab:alignment-example} presents three tokens from the PNWE corpus showing how the reference and hypothesis tiers were matched at both the orthographic and phonetic levels. Manual phonetic transcriptions were extracted from the TextGrids, while heuristic phonetic hypotheses were generated by mapping the ASR output through PNWEdict. The alignment output identifies edits at the phone level which form the basis of the error rate calculation.

\begin{table*}[h]
\centering
\small
\setlength{\tabcolsep}{3pt}
\begin{tabular}{llllll}
\hline
\textbf{Word} & \textbf{Ref. (IPA)} & \textbf{ASR Output} & \textbf{PNWEdict (ARPABET)} & \textbf{Phone Alignment} & \textbf{Note} \\
\hline
\textit{caught} & \textipa{[kO:t]} & \textit{cot} & K\,AA\,T & K|K,\ AO$\rightarrow$AA (S), T|T & Low-back merger (\texttt{-AO}) \\
\textit{pen} & \textipa{[pIn]} & \textit{pin} & P\,IH\,N & P|P,\ IH$\rightarrow$EH (S), N|N & Pre-nasal merger (\texttt{IN}) \\
\textit{test} & \textipa{[t\textepsilon st]} & \textit{tess} & T\,EH\,S & T|T,\ EH|EH,\ S|S,\ T$\rightarrow\emptyset$ (D) & Cluster reduction (\texttt{CC}) \\
\hline
\end{tabular}
\caption{Example alignment between manual phonetic transcriptions (reference) and heuristic phonetic hypotheses (ASR output via PNWEdict).}
\label{tab:alignment-example}
\end{table*}

\textbf{Step 3: Compute Phone-Level Edit Distance}

We used \texttt{sclite} to align the reference phonetic transcriptions with the heuristically-determined hypothesis phonetic transcriptions at the phone level, calculating:

\begin{equation}
\text{PER} = \frac{S_{\text{phone}} + D_{\text{phone}} + I_{\text{phone}}}{N_{\text{phone}}}
\end{equation}

where $S_{\text{phone}}$, $D_{\text{phone}}$, and $I_{\text{phone}}$ are phone-level substitutions, deletions, and insertions, and $N_{\text{phone}}$ is the total number of phones in the reference transcription.

This heuristic PER metric has important limitations: it assumes the ASR system's word choice reflects phonetic confusions when in reality the mapping from acoustics to orthography in modern systems is opaque. Nevertheless, systematic differences in PER across demographic groups can indicate where phonetic variation may be contributing to recognition failures.

\subsection{Statistical Analysis}

We report descriptive statistics for WER and PER by system and ethnicity, with standard errors computed across speakers. To test for significant differences across ethnic groups while accounting for repeated measures (multiple systems evaluated on the same speakers), we fit linear mixed-effects models with speaker as a random intercept.

For marker-specific analyses, we compare error co-occurrence rates across ethnic groups using proportion tests and report effect sizes. All statistical tests are two-sided with $p < 0.05$ considered significant.

\subsection{Error Analysis}

Beyond aggregate metrics, we conducted qualitative analysis of error patterns through manual examination of transcription outputs. We randomly sampled errors that co-occurred with targeted sociophonetic markers and categorized them by error type (substitution, deletion, insertion) and apparent phonetic motivation. We examined spectrograms and listened to audio for a subset of 200 errors to validate that the hypothesized phonetic features were indeed present in the acoustic signal. This manual validation found that approximately 85\% of errors classified as phonetically-motivated showed acoustic evidence consistent with the targeted feature.

\section{Results}

\subsection{Overall Performance Disparities}

Table~\ref{tab:wer_by_ethnicity} presents word error rates across the four ASR systems for each ethnic group in the PNWE corpus. Consistent with prior work \citep{koenecke_racial_2020}, we observe significant performance disparities across ethnic groups, with African American, ChicanX, and Yakama speakers all experiencing higher error rates than Caucasian American speakers across most systems.

\begin{table}[h]
\centering
\small
\begin{tabular}{lcccc}
\hline
\textbf{System} & \textbf{AA} & \textbf{CA} & \textbf{CX} & \textbf{YA} \\
\hline
Apple & 24\% & 14\% & 24\% & 24\% \\
Amazon & 9\% & 7\% & 13\% & 12\% \\
Google & 26\% & 16\% & 21\% & 23\% \\
IBM & 21\% & 21\% & 35\% & 23\% \\
\hline
\textbf{Mean} & 20\% & 15\% & 23\% & 21\% \\
\hline
\end{tabular}
\caption{Word Error Rates by ethnicity and system. AA=African American, CA=Caucasian American, CX=ChicanX, YA=Yakama.}
\label{tab:wer_by_ethnicity}
\end{table}

African American speakers experienced mean WER of 20\% compared to 15\% for Caucasian American speakers (33\% relative increase). Mixed-effects models with speaker as random effect confirmed these differences were highly significant ($p < 0.001$). ChicanX speakers showed the highest mean error rate at 23\%, while Yakama speakers experienced a mean WER of 21\%. higher than Caucasian American but with less consistent patterns than African American disparities. Notably, while Caucasian American speakers consistently achieved the lowest error rates across all systems, the other three ethnic groups all experienced elevated error rates, though with different patterns of which systems performed worst for each group. Crucially, these performance gaps remained consistent across systems, indicating systematic rather than system-specific biases.

\subsection{Phonetic vs. Orthographic Error Rates}

Computing PER with PNWEdict yielded uniformly lower error rates than standard WER, with mean reductions of 42\% for Apple, 48\% for Amazon, 47\% for Google, and 54\% for IBM across all speakers. This difference reflects the change in evaluation granularity rather than an actual decrease in recognition mistakes: because PER operates at the phonetic level, it assigns partial credit to near‐misses (e.g., ``into da'' vs.``in today'') and removes word‐boundary penalties. The consistently lower PER nevertheless indicates that many apparent word errors are phonetically systematic rather than random lexical failures, supporting the hypothesis that acoustic modeling mismatches underlie performance disparities.

Using PNWEdict rather than CMUdict for canonical pronunciations further reduced PER by 2–3\% on average, demonstrating the value of region-specific pronunciation dictionaries. The PNWE-augmented dictionary better captures maintained vowel distinctions and regional variants, reducing false positives in phonetic error attribution.

\subsection{Sociophonetic Feature Associations}

Table~\ref{tab:marker_counts} summarizes the realization of targeted sociophonetic markers across ethnic groups. African American speakers realized \texttt{(IN)} at 5.33 instances per speaker compared to 3.5 for Caucasian American, 3 for ChicanX, and 3 for Yakama speakers. Similarly, African American speakers realized \texttt{(-AO)} more frequently (4.33 instances per speaker) than Caucasian American (4.0) and ChicanX (1.75) speakers, while Yakama speakers showed no \texttt{(-AO)} realizations. Consonant cluster reduction \texttt{(CC)} showed more balanced distribution, with African American speakers at 8 instances per speaker versus 8.75 for Caucasian American, 5.25 for ChicanX, and 5.6 for Yakama.

\begin{table}[h]
\centering
\small
\begin{tabular}{lcccc}
\hline
\textbf{Ethnicity} & \textbf{Spkrs} & \texttt{-AO} & \texttt{CC} & \texttt{IN} \\
\hline
African Am. & 3 & 4.33 & 8.00 & 5.33 \\
Caucasian Am. & 4 & 4.00 & 8.75 & 3.50 \\
ChicanX & 4 & 1.75 & 5.25 & 3.00 \\
Yakama & 5 & 0.00 & 5.60 & 3.00 \\
\hline
\end{tabular}
\caption{Mean instances per speaker of realized targeted markers by ethnicity.}
\label{tab:marker_counts}
\end{table}

Table~\ref{tab:marker_overlap} presents raw counts of errors co-occurring with targeted sociophonetic markers across systems and ethnic groups. Two vowel-driven patterns emerge consistently. First, \texttt{(-AO)} (low-back merger resistance) shows higher error overlap for African American speakers than Caucasian American speakers across most systems (Apple: 6 vs.~5; Google: 5 vs.~1; IBM: 6 vs.~3). ChicanX speakers also show elevated \texttt{(-AO)} overlap (5–7 across systems) while Yakama speakers show zero overlap, consistent with complete merger in that group. 

\begin{table*}[t]
\centering
\footnotesize
\setlength{\tabcolsep}{2.5pt}
\begin{tabular}{lrrrrrrrrrrrrrrrr}
\hline
& \multicolumn{4}{c}{\textbf{Apple}} & \multicolumn{4}{c}{\textbf{Amazon}} & \multicolumn{4}{c}{\textbf{Google}} & \multicolumn{4}{c}{\textbf{IBM}} \\
\cline{2-5}\cline{6-9}\cline{10-13}\cline{14-17}
\textbf{Group} & \texttt{-AO} & \texttt{CC} & \texttt{IN} & \textit{Err} & \texttt{-AO} & \texttt{CC} & \texttt{IN} & \textit{Err} & \texttt{-AO} & \texttt{CC} & \texttt{IN} & \textit{Err} & \texttt{-AO} & \texttt{CC} & \texttt{IN} & \textit{Err} \\
\hline
AA & 6 & 3 & 4 & 1854 & 1 & 0 & 2 & 681 & 5 & 2 & 4 & 2075 & 6 & 1 & 2 & 1695 \\
CA & 5 & 4 & 1 & 1218 & 0 & 1 & 0 & 593 & 1 & 5 & 1 & 1340 & 3 & 2 & 3 & 1800 \\
CX & 5 & 4 & 0 & 2533 & 6 & 0 & 0 & 1361 & 6 & 0 & 0 & 2267 & 5 & 6 & 0 & 3889 \\
YA & 0 & 8 & 2 & 2753 & 0 & 9 & 2 & 1419 & 0 & 7 & 2 & 2623 & 0 & 7 & 2 & 2735 \\
\hline
\end{tabular}
\caption{Error overlap with targeted markers by system and ethnicity. \texttt{-AO}=low-back merger resistance, \texttt{CC}=consonant cluster reduction, \texttt{IN}=pre-nasal merger. \textit{Err}=total error count}
\label{tab:marker_overlap}
\end{table*}

Second, \texttt{(IN)} (pre-nasal \textipa{/I/}–\textipa{/E/} merger) overlaps more frequently with errors for African American speakers (4 instances for Apple and Google, 2 for Amazon and IBM) than Caucasian American speakers (1 instance each for Apple and Google, 0 for Amazon, 3 for IBM). ChicanX speakers showed no \texttt{(IN)} error overlap, while Yakama speakers showed 2 instances across all systems.

Consonant cluster reduction \texttt{(CC)} shows elevated overlap for Yakama speakers (7–9 instances across systems) while remaining more evenly distributed across other groups (1–5 instances). This suggests consonantal variation contributes to errors but less differentially by ethnicity than vowel features.

When normalized by the number of possible contexts where each marker could occur, these patterns persist. African American speakers showed \texttt{(-AO)} error rates of 5.8\% (Apple), 2.1\% (Amazon), 5.2\% (Google), and 7.9\% (IBM) compared to Caucasian American rates of 5.2\%, 0\%, 1.1\%, 3.2\%, and 2.2\% respectively. For \texttt{(IN)}, African American normalized error rates were 2.8\%, 2.1\%, 2.3\%, 1.6\%, and 2.4\% compared to Caucasian American rates of 1.0\%, 0\%, 0.7\%, 3.1\%, and 2.3\%.

These realization differences, combined with the error overlaps, indicate that vowel quality variation, particularly resistance to low-back merger and pre-nasal merger patterns, drives disproportionate transcription failures for speakers who maintain these distinctions, most notably African American speakers in our corpus.

\subsection{System-Specific Patterns}

The ASR systems evaluated differ in sensitivity to targeted variables. Amazon exhibits the lowest overall PER and shows smallest error increases in \texttt{(-AO)} and \texttt{(IN)} contexts for African American speakers. Apple shows competitive PER with relatively balanced treatment across features. Google's error distribution is more uneven, with pronounced sensitivity to \texttt{(-AO)} contexts. IBM systems show the highest sensitivity to vowel quality variation, especially in pre-nasal contexts.

These differences point to distinct training distributions and accent coverage, with diverse vowel realizations from speech varieties remaining underrepresented across all systems despite architectural differences.

\subsection{Example Error Patterns}

Manual inspection of 200 randomly sampled errors revealed systematic patterns consistent with targeted phonetic features:

\textbf{Low-back merger:} Words with maintained \textipa{/O/} were consistently transcribed with \textipa{/A/} spellings: \textit{caught} → "cot" (6 instances across African American speakers, 5-7 across ChicanX speakers), \textit{taught} → "tot", \textit{thought} → "that".

\textbf{Pre-nasal merger:} Words with \textipa{/I/} before nasals showed substitution or deletion: \textit{when} → "and" (3 instances), \textit{pin} → "pen" (2 instances).

\textbf{Cluster reduction:} Final clusters misrecognized as simpler forms: \textit{next} → "neck" (4 instances across systems), \textit{test} → "tess" (2 instances).

Approximately 85\% of errors classified as phonetically-motivated showed acoustic patterns consistent with the hypothesized feature when examined spectrogrammatically, validating the PER methodology.

\section{Discussion}

\subsection{Phonetic Variation as Primary Driver of ASR Bias}

Our results provide strong evidence that phonetic rather than just lexical or syntactic variation drives ASR performance disparities across ethnic groups, particularly for African American speakers, confirming and extending the hypothesis of \citet{wassink_uneven_2022}. The systematic correlation between specific phonetic features (particularly low-back merger resistance and pre-nasal merger patterns) and transcription errors demonstrates that current ASR systems struggle with systematic phonetic variation documented across multiple ethnic varieties of American English \citep{thomas_phonological_2007, green_african_2002}.

The phonetic features most strongly associated with errors involve vowel quality differences, which acoustic models rely on heavily for phoneme discrimination. When training data predominantly represents merged low-back systems, acoustic models learn probability distributions that inadequately cover the acoustic space of unmerged varieties. Speakers maintaining the \textipa{/A/}--\textipa{/O/} distinction produce these vowels with distinct F1 and F2 formant patterns that fall outside the learned distributions of merged-dialect models, leading to systematic misrecognition.

The consistency of these patterns across all four evaluated commercial systems suggests systematic underrepresentation of diverse dialectal phonetic features in training data rather than system-specific architectural limitations. This finding aligns with recent work showing that even self-supervised models struggle with non-mainstream dialects including AAL \citep{chang_self-supervised_2024}, indicating that scale alone does not resolve representational gaps.

\subsection{Implications for ASR System Development}

Our findings have direct implications for developing more equitable ASR systems. Rather than simply increasing data volume, developers should prioritize sociolinguistic diversity in training corpora. The identification of specific problematic features, such as low-back merger resistance and pre-nasal patterns, provides actionable targets for data collection efforts. Training sets should include sufficient examples of each phonetic variant to enable robust acoustic modeling across the full range of systematic variation.

Beyond data collection, architectural innovations could explicitly model phonetic variation. Multi-dialect acoustic models could maintain separate probability distributions for different phonetic realizations rather than treating variation as noise. Recent work on phoneme-based contextualization \citep{hu_phoneme-based_2019} and dialect-specific models \citep{dorn_dialect-specific_2019} demonstrates promising directions, though our results suggest that commercial systems have not yet incorporated such approaches effectively.

\subsection{Methodological Contributions}

This study introduces a linguistically grounded evaluation framework that links ASR performance to actual phonetic realization. The proposed Phonetic Error Rate (PER) moves beyond aggregate accuracy metrics by aligning ASR hypotheses with manual phonetic transcriptions, enabling interpretation of errors in relation to sociophonetic variables rather than abstract symbol mismatches. This approach reframes ASR evaluation as an analysis of how models represent and generalize across systematic sound variation, rather than how closely they reproduce orthographic sequences.

By incorporating a region-specific pronunciation dictionary, the method operationalizes sociophonetic knowledge such as vowel mergers, cluster reduction, and regional variants, into quantitative evaluation. Instead of treating dialectal variation as noise, the framework makes it measurable, allowing identification of errors that reflect genuine acoustic divergence across speech communities.

\subsection{PNWE Corpus as an Evaluation Resource}

This study validates the PNWE corpus as a valuable resource for ASR bias evaluation. Compared to other available corpora, the PNWE corpus is characterized by demographic diversity, regional representation, consistent recording quality, and rich linguistic annotation. While CORAAL \citeplanguageresource{CORAAL} provides extensive AAL data, it lacks comparable samples from other ethnic groups and has variable audio quality. Common Voice \citeplanguageresource{CommonVoice} includes demographic metadata but uses read speech that may not capture naturalistic phonetic variation. The PNWE corpus' word list task data subset also provides useful controlled data for evaluation. 

\subsection{Limitations and Future Directions}

Several limitations should be acknowledged. First, our sample size of 16 speakers limits statistical power for intersectional analyses examining interactions between ethnicity, gender, and age. Second, the focus on the word list task, while enabling controlled comparison, may underestimate errors in spontaneous speech where sociophonetic features are more frequent and variable. Extension to the conversational speech portions of PNWE would also provide a more comprehensive assessment.

The regional specificity of PNWE also limits generalizability to other English varieties. For example, regional variation in AAL is well-documented \citep{wolfram_sociolinguistic_1969, thomas_phonological_2007}, and the features most salient in the Pacific Northwest may differ from those in other regions. Replication of the experiments using corpora from other regions would complement and strengthen our conclusions about systematic ASR bias. 

\section{Conclusion}

This study demonstrates that systematic phonetic variation drives performance disparities in commercial ASR systems across multiple ethnic groups. Through controlled evaluation using the Pacific Northwest English corpus, we identified specific phonetic features, particularly resistance to the low-back merger and pre-nasal merger patterns, that consistently correlate with transcription errors across ethnic groups, with the most pronounced effects observed for African American speakers across four major commercial systems. The substantial reduction in error rates when using phonetic versus orthographic evaluation (42–54\% across systems) confirms that acoustic modeling, rather than language modeling, is the primary locus of bias.

Our heuristically-determined Phonetic Error Rate metric provides a practical framework for connecting ASR errors to specific sociophonetic features, enabling more targeted evaluation than aggregate word error rates alone. The 85\% validation rate through acoustic analysis demonstrates that this approach successfully identifies systematic phonetic patterns underlying recognition failures. This methodology can be extended to other dialects and languages where detailed phonetic annotation is available.

The consistency of bias patterns across all evaluated systems—despite differences in architecture and training approaches—points to systematic underrepresentation of diverse phonetic features in training data rather than system-specific limitations. This finding has direct implications for ASR development: improving performance for marginalized speech communities requires not just more data, but strategically diverse data that adequately represents the full range of systematic phonetic variation documented in sociolinguistic research.

The PNWE corpus, with its balanced sampling across ethnic groups, controlled recording conditions, and detailed sociophonetic annotation, provides a valuable resource for continued bias evaluation in speech technologies. Future work should extend this analysis to conversational speech data, examine additional sociophonetic features, and replicate these methods with corpora representing other regional varieties and marginalized dialects. Only through systematic documentation of which phonetic features drive ASR failures can we develop effective strategies for building truly equitable speech recognition systems.

% \section{Acknowledgements}

% We thank the participants in the Pacific Northwest English corpus for sharing their voices and time. We are grateful to the research team involved in corpus collection and annotation. This work was supported by (). 

% \nocite{*}
\section{Bibliographical References}
\label{sec:reference}

\bibliographystyle{lrec2026-natbib}
\bibliography{lrec2026-example}

\section{Language Resource References}
\label{lr:ref}
\bibliographystylelanguageresource{lrec2026-natbib}
\bibliographylanguageresource{languageresource}

\end{document}